\documentclass[11pt]{article}

\usepackage[a4paper,margin=1in]{geometry}
\usepackage[T1]{fontenc}
\usepackage[utf8]{inputenc}   
\usepackage{lmodern}          
\usepackage{microtype}

\usepackage{amsmath,amssymb,amsthm,mathtools}

\usepackage{graphicx}
\usepackage{booktabs}
\usepackage{tabularx}
\usepackage{longtable}
\usepackage{multirow}
\usepackage{siunitx}

\usepackage{float}

\usepackage[ruled,vlined]{algorithm2e} 
\SetKwInput{KwParams}{Parameters} 

\usepackage[numbers,sort&compress]{natbib}

\usepackage[hidelinks]{hyperref}

\usepackage{caption}
\usepackage{graphicx}
\usepackage{subcaption}
\usepackage{tikz}
\usetikzlibrary{positioning}
\usepackage{pgfplots}
\pgfplotsset{compat=1.18}

\usepackage[hidelinks]{hyperref}

\newtheorem{problem}{Problem}
\newtheorem{remark}{Remark}

\usepackage{tabularx}
\usepackage{caption}
\usepackage{tikz}
\usetikzlibrary{positioning}
\usepackage{pgfplots}
\pgfplotsset{compat=1.18}
\usepackage{booktabs}
\usepackage{longtable}
\usepackage{siunitx} 
\usepackage{multirow}
\usepackage{graphicx}
\usepackage{hyperref}

\usepackage{mathtools}

\title{An Enhanced Focal Loss Function to Mitigate Class Imbalance in  Auto Insurance Fraud Detection with Explainable AI}
\newcommand{\TMUaffil}{Department of Mathematics, Toronto Metropolitan University, Toronto, Canada. \\ Corresponding author: asante.gyamerah@torontomu.ca}

\author{%
	Francis Boabang\thanks{\TMUaffil} \and
	Samuel Asante Gyamerah\footnotemark[1] %
}
\date{} 

\begin{document}
	\maketitle
	
	\begin{abstract}
		\noindent
		Detecting fraudulent auto-insurance claims remains a challenging classification problem, largely due to the extreme imbalance between legitimate and fraudulent cases. Standard learning algorithms tend to overfit to the majority class, resulting in poor detection of economically significant minority events. This paper proposes a structured three-stage training framework that integrates a convex surrogate of focal loss for stable initialization, a controlled non-convex intermediate loss to improve feature discrimination, and the standard focal loss to refine minority-class sensitivity. We derive conditions under which the surrogate retains convexity in the prediction space and show how this facilitates more reliable optimization when combined with deep sequential models. Using a proprietary auto-insurance dataset, the proposed method improves minority-class F1-scores and AUC relative to conventional focal-loss training and resampling baselines. The approach also provides interpretable feature-attribution patterns through SHAP analysis, offering transparency for actuarial and fraud-analytics applications. \url{https://github.com/boabangf/Efficient-Focal-Loss-Function-for-Insurance-Fraud-Prediction}
	\end{abstract}
	\noindent\textbf{Keywords:} automobile insurance fraud prediction; class imbalance; explainable AI; multistage focal loss function; machine learning.

\section{Introduction}
\label{intro}
\noindent
Fraudulent auto-insurance claims continue to impose substantial financial and operational costs on insurers, distort the equity of premium setting, and weaken broader market stability. Although predictive analytics has advanced considerably, the reliable identification of fraudulent claims remains difficult because such cases constitute only a small proportion of the overall portfolio. This severe class imbalance encourages conventional classifiers to optimize primarily for overall accuracy, often at the expense of accurately detecting rare but financially significant events. For actuarial practice where missed fraud directly affects loss ratios and pricing adequacy, methods capable of capturing sequential claim patterns and learning effectively from limited minority samples are particularly valuable. 
\\
[2mm]
Auto-insurance fraud detection is an essential yet difficult problem for insurers, largely due to the extreme imbalance between genuine and fraudulent claims \citep{ding2025automobile}. In nearly all operational datasets, legitimate claims vastly outnumber fraudulent ones \citep{johnsonhandling}, causing standard learning algorithms to optimize for overall accuracy rather than identifying the rare but economically consequential minority class. As a result, many models fail to detect fraudulent behaviour or produce excessive false alarms, outcomes that directly increase claim costs and erode pricing fairness. Addressing this imbalance is therefore important for building reliable fraud analytics systems. To mitigate this imbalance, different studies have commonly employ resampling, cost-sensitive learning, ensemble methods, and modified loss functions such as focal loss \citep{hanafy2021using}. Focal loss, in particular, has gained substantial attention for its ability to down-weight easy-to-classify majority examples and allocate learning emphasis to difficult minority cases \citep{ross2017focal}. However, the standard focal-loss formulation employs a fixed focusing parameter, $\gamma$, throughout training. This rigidity can lead to unstable gradients early on, prematurely discount easy observations, and fail to adapt to the evolving difficulty distribution as model parameters change.
\\
[2mm]
A number of adaptive loss formulations have been suggested in recent years, each attempting to address the instability introduced by the standard focal loss. CurricularFace \citep{huang2020curricularface} frames the problem in terms of a learning curriculum, encouraging the network to concentrate on progressively more difficult cases as training unfolds. LDAM \citep{cao2019ldam} takes a different route by adjusting class margins in proportion to observed class frequencies. These methods, however, do not transfer seamlessly to fraud-analytics contexts: CurricularFace was developed for high-dimensional face-recognition embeddings, and LDAM presumes that class proportions are reliably known and stable, an assumption that seldom holds in operational insurance data. AdaFocal \citep{ghosh2022adafocal} moves closer to what is needed by updating the focusing parameter $\gamma$ through feedback from a validation set, drawing on the broader literature showing that well-calibrated models can improve decision reliability \citep{mukhoti2020calibrating}. Even so, AdaFocal remains dependent on periodic validation-driven adjustments and binning schemes, which complicate its use in strictly offline or batch-learning environments common in insurance practice.
\\
[2mm]
One difficulty that continues to arise across adaptive focal-type losses is the pronounced non-convexity they introduce into the optimisation landscape. In practice, this often leads to unstable training behaviour, with outcomes that depend strongly on the initial parameter values or drift toward unwanted stationary points. These challenges become more acute when the loss is paired with deep sequence-based architectures, which already carry well-known issues related to vanishing gradients and irregular convergence patterns. As a result, the focal-loss framework, while conceptually appealing for imbalanced classification, can be susceptible to inconsistent optimisation trajectories in real actuarial applications. This motivates the development of more structured mechanisms that preserve the discriminative advantages of focal loss while providing a more reliable path through the optimisation process.
\\
[2mm]
In this paper, we propose a three-stage focal-loss function that addresses these challenges within an end-to-end machine-learning pipeline for auto-insurance fraud detection. Unlike other adaptive focal loss approaches \cite{cao2019ldam,huang2020curricularface, ghosh2022adafocal} that can struggle with local optima due to their non-convex nature, the proposed three-stage design incorporates iterative refinements that guide the optimization process more effectively, thereby enhancing convergence stability. The framework also includes robust data SMOTE \cite{chawla2002smote}, an oversampling technique to partially handle the class imbalance issue. Using a proprietary auto-insurance claims dataset, we demonstrate that the proposed method consistently outperforms traditional focal-loss training and representative baselines. Additionally, the integration of feature-importance and correlation-analysis techniques provides interpretability aligned with operational fraud-investigation needs. 

\section{Related Work}
\label{sec:Related Work}
\noindent
Insurance fraud detection has attracted considerable academic and industry attention due to its implications for financial stability, operational efficiency, and equitable premium pricing. Research in this field typically follows three broad strands: (i) the development of statistical and machine-learning methods for analyzing claims data; (ii) the design of techniques to address class imbalance, which is a defining characteristic of fraud datasets; and (iii) the use of explainable artificial intelligence to support transparency and operational adoption. This section reviews relevant contributions across these domains, with a particular focus on machine-learning approaches for auto-insurance fraud detection and recent advances in model interpretability.
\\
[2mm]
Auto-insurance fraud detection has long been recognized as a critical application of predictive modelling because fraudulent activity imposes substantial financial losses on insurers and ultimately affects policyholders through increased premiums. Early studies employed classical techniques such as logistic regression and decision trees, but these models often struggled to capture complex fraud patterns, particularly under severe class imbalance \citep{ngai2011application, caruana2021automobile}. To improve predictive performance, numerous studies have turned to ensemble methods, which generally offer superior robustness in structured tabular data. Random Forests and XGBoost, in particular, have demonstrated strong performance in identifying suspicious claims and have become widely used in fraud analytics pipelines \citep{feitosa2021study, subudhi2020use, rubaidi2023vehicle}. A parallel line of work highlights the importance of hyperparameter optimisation. While exhaustive search strategies such as grid search or random search can be computationally demanding, metaheuristic approaches, including Genetic Algorithms (GA), Simulated Annealing (SA), and Particle Swarm Optimization (PSO) have gained prominence for their efficiency and flexibility \citep{tayebi2021hyperparameter, dalal2022predicting}. PSO, in particular, has been shown to enhance the performance of XGBoost and related models in auto-insurance fraud contexts \citep{ding2025automobile}.
\\
[2mm]
Given the extreme imbalance between legitimate and fraudulent claims, many studies make use of resampling techniques, including SMOTE, ADASYN, and Random Undersampling (RUS), to reduce bias in model training \citep{hanafy2021using}. Integrating these methods with ensemble learners and robust optimisation strategies has been shown to improve both precision and recall \citep{mrozek2020efficient}. \citep{gheysarbeigi2025ensemble} applied a Binary Quantum-Based Avian Navigation Optimizer (BQANA) to tune ensemble classifiers (SVM, RF, and XGBoost) in conjunction with undersampling, achieving improved accuracy and recall over conventional tuning strategies. Another recent study \citep{khalil2024machine} combined multiple imputation, resampling, and ensemble learning to assess the combined effect of these components on fraud detection. While these studies demonstrate the value of resampling and ensemble strategies, many rely heavily on synthetic datasets and do not incorporate loss-function-based optimization strategies, such as focal loss, which are specifically designed to handle extreme imbalance. To address this gap, \cite{wongpanti2024enhancing}incorporated Focal Loss into a one-dimensional convolutional neural network (1D-CNN) for auto-insurance fraud detection. 
Focal Loss, originally introduced for dense object detection, reduces the influence of well-classified (majority-class) samples and concentrates learning on difficult minority-class observations. Although this approach showed improvements in fraud-detection accuracy, it also highlighted a central challenge: the non-convex nature of focal loss can complicate optimisation and increase the likelihood of convergence to suboptimal local minima, particularly in highly imbalanced settings.

\subsection{Recent Developments and Interpretability}
\noindent
A growing segment of recent research emphasizes not only improving predictive accuracy in fraud detection but also enhancing model interpretability. \cite{yankol2022value} employed Local Interpretable Model-Agnostic Explanations (LIME) to interpret automobile insurance fraud models and assess the relative influence of the features. \cite{debener2023detecting} conducted a comprehensive comparison of supervised and unsupervised methods using a large proprietary insurance dataset. \cite{maiano2023deep} presented an automatic end-to-end system for identifying fraud attempts in the insurance sector. \cite{nordin2024predicting} demonstrated that AdaBoost can substantially improve the predictive strength of decision-tree models. Many of these studies rely on cooperative Shapley-value frameworks to attribute predictions to features. We therefore review both cooperative and non-cooperative Shapley-based approaches to situate our work within the broader interpretability landscape.

\section{Formulation}
\label{sec:Formulation}
\section*{Problem Definition}
\begin{problem}
	\noindent
	Let  $\mathcal{D} = \{(x_i, y_i)\}_{i=1}^N$ denote a labelled dataset comprising \(N\) temporal samples, where each input sequence  
	$x_i \in \mathbb{R}^{T \times d}$
	represents a time series of length \(T\) with \(d\) features at each time step, and the corresponding label  $y_i \in \{0, 1\}$ indicates whether the case is fraudulent (\(y_i = 1\)) or non-fraudulent (\(y_i = 0\)).
	In a fraud detection system, a major problem is class imbalance. The dataset is usually highly skewed, with legitimate (non-fraudulent) cases becoming more than the fraudulent ones. Formally,
	\[
	|\{y_i = 1\}| \ll |\{y_i = 0\}|,
	\]
	resulting in biased learning behaviour where conventional loss functions favour the majority class. Consequently, models may achieve high overall accuracy while not detecting correctly the minority fraudulent instances.
\end{problem}

\subsection*{LSTM Model Architecture}
\noindent
We consider a Long Short-Term Memory (LSTM) network $f_{\theta}$, parameterised by weights $\theta$, which maps an input sequence $x_i$ to a probability score:
\begin{equation}
	p_i = f_{\theta}(x_i) \in [0, 1].
\end{equation}
The output $p_i$ represents the model’s predicted probability that sequence $x_i$ belongs to the positive (fraudulent) class, that is, $y_i = 1$.

\section*{Focal Loss Function: Convex and Non-Convex Formulations}
\noindent
Focal loss is a modification of the standard cross-entropy loss designed to address class imbalance by placing greater emphasis on hard-to-classify examples \cite{ross2017focal}. It introduces a modulating factor $(1 - p_t)^\gamma$ into the cross-entropy loss, where $p_t$ denotes the model’s estimated probability for the true class.

\subsection*{Non-Convex Focal Loss Function}
\noindent
The original focal loss \cite{ross2017focal} is typically non-convex and is defined as
\begin{equation}
	\text{FL}_{\text{non-convex}}(p_t)
	= -\alpha_t (1 - p_t)^\gamma \log(p_t),
\end{equation}
where
\begin{itemize}
	\item $p_t = p$ if $y = 1$ and $p_t = 1 - p$ if $y = 0$,
	\item $\alpha_t$ is a class-weighting factor used to compensate for imbalance,
	\item $\gamma \geq 0$ is the focusing parameter.
\end{itemize}
For $\gamma > 0$, the multiplicative factor $(1 - p_t)^\gamma$ alters the curvature of the loss in a way that leads to a non-convex landscape in $p_t$ and, more importantly, in the model parameters $\theta$ when combined with a non-linear network.

\subsection*{Convex Approximation of Focal Loss}
\noindent
To obtain a more tractable loss for the initial phase of training, we consider a convex surrogate inspired by \cite{yu2016convex, zhang2004statistical, qaraei2021convex}:
\begin{equation}
	\text{FL}_{\text{convex}}(p_t)
	= -\alpha_t (1 - \gamma p_t)\log(p_t),
\end{equation}
where $p_t \in (0,1)$ is the predicted probability for the true class, $\gamma > 0$ is the focusing parameter, and $\alpha_t$ is a fixed class-balancing coefficient. The associated weight can be written as
\begin{equation}
	w_t = -\alpha_t (1 - p_t)^{\gamma}.
\end{equation}
\noindent
In our convex surrogate formulation, $w_t$ is treated as fixed, and a convex activation function such as \texttt{softplus} is used in the final layer. This decoupling is important: when $w_t$ is allowed to vary with $p_t$, the dependence between the weight and the prediction term introduces additional curvature that can destroy convexity. By fixing $w_t$ and using a convex activation in the logit space, we can analyse the loss purely in terms of convexity conditions on $p_t$.
\\
Consider
\[
f(p) = - (1 - \gamma p)\log(p),
\]
with $\alpha_t$ treated as a constant that does not affect convexity. The first derivative is
\begin{align}
	f'(p)
	&= \frac{d}{dp}\left[-(1 - \gamma p)\log(p)\right] \\
	&= \gamma \log(p) - \frac{1 - \gamma p}{p},
\end{align}
and the second derivative is
\begin{align}
	f''(p)
	&= \frac{d}{dp}\left[\gamma \log(p) - \frac{1 - \gamma p}{p}\right] \\
	&= \frac{\gamma}{p} + \frac{1}{p^2}.
\end{align}
Since $p \in (0,1)$ and $\gamma > 0$, both $\frac{\gamma}{p}$ and $\frac{1}{p^2}$ are strictly positive, hence
\begin{equation}
	f''(p) > 0 \quad \forall p \in (0,1),
\end{equation}
which implies that $f(p)$ is strictly convex on $(0,1)$.
\\
The convexity result therefore holds under the assumption that $w_t$ is fixed and that we analyse the loss as a function of $p_t$ (or equivalently the logit $z$ through a convex link). If $w_t$ were a function of $p_t$, additional terms involving $\frac{d w_t}{d p_t}$ would arise, introducing components that may break convexity. By employing a convex activation and fixing $w_t$, the loss landscape in the probability/logit space remains well behaved, which is precisely what we require for a stable first-stage optimisation.
\\
Furthermore, the condition $1 - \gamma p_t > 0$ (for all relevant $p_t$) can be enforced by an appropriate choice of $\gamma$ so that the sign of the loss and its curvature do not invert.
\\
Under these assumptions, we have
\begin{equation}
	f''(p) = \frac{\gamma}{p} + \frac{1}{p^2} > 0,
\end{equation}
confirming that the proposed focal surrogate is strictly convex in $p \in (0,1)$. This provides a mathematically grounded basis for employing a convex-stage optimisation in our three-stage (convex–to–non-convex) training strategy, yielding a stable initialisation for subsequent adaptive refinement.
\\
In practice, the convex variant used in Stage~1 is implemented as
\[
\mathcal{L}(y,z)
= \alpha (1 - p_t)^{\gamma}
\Big[\, y \cdot \mathrm{softplus}(-z)
+ (1-y)\cdot \mathrm{softplus}(z) \Big],
\]
where $z$ denotes the model logit, $y \in \{0,1\}$ is the label, and
$p_t = p$ for $y=1$ and $p_t = 1-p$ for $y=0$. When $\gamma = 0$, the multiplicative focal term disappears:
\[
(1 - p_t)^{\gamma} = 1,
\]
and the loss reduces to the classical softplus–logistic form:
\[
\mathcal{L}(y,z)
= y \cdot \mathrm{softplus}(-z)
+ (1-y)\cdot \mathrm{softplus}(z).
\]
The softplus function,
\[
\mathrm{softplus}(z) = \log(1 + e^{z}),
\]
is smooth and strictly convex, implying that $\mathcal{L}(y,z)$ is convex in the logit $z$. Although the overall network remains non-convex in its parameters, the first stage is convex in terms of the output logits, and thus acts as a convex warm start that stabilises early training before transitioning to the later non-convex phases.

\begin{algorithm}[H]
	\small
	\caption{Convex Focal Loss with Softplus Logistic Link}
	\label{alg:convex_focal}
	
	\KwIn{True labels $y \in \{0,1\}$, predicted logits $z \in \mathbb{R}$}
	\KwParams{Focal parameter $\gamma=0$, class weight $\alpha\in(0,1)$}
	\KwOut{Scalar loss $\mathcal{L}_{\text{batch}}$}
	
	\textbf{Convex Softplus logistic surrogates:}\;
	\[
	\ell^{+}(z)=\log(1+e^{-z}),\qquad \ell^{-}(z)=\log(1+e^{z})
	\]
	
	\textbf{Convex probability surrogate:}\;
	\[
	p=\exp(-\ell^{+}(z))
	\]
	
	\textbf{Class-conditioned probability:}\;
	\[
	p_t = y\,p + (1-y)(1-p)
	\]
	
	\textbf{Convex focal weight approximation:}\;
	\[
	w_t=\exp(-\gamma p_t)
	\]
	
	\textbf{Base convex logistic loss:}\;
	\[
	\ell_{\text{base}} = y\,\ell^{+}(z) + (1-y)\,\ell^{-}(z)
	\]
	
	\textbf{Final convex focal loss:}\;
	\[
	\mathcal{L} = \alpha\, w_t\, \ell_{\text{base}}
	\]
	
	\textbf{Batch reduction:}\;
	\[
	\mathcal{L}_{\text{batch}} = \frac{1}{N}\sum_{i=1}^{N}\mathcal{L}_i
	\]
	
	\Return{$\mathcal{L}_{\text{batch}}$}\;
\end{algorithm}

\begin{remark}
	In Stage~1 of the model (see Algorithm~\ref{alg:class_imbalance}), all dense layers are \emph{convex-aware}: each consists of a linear transformation followed by a convex, non-decreasing activation such as ReLU or softplus. By standard results from convex analysis, mappings of the form $\mathrm{ReLU}(Wx+b)$ and $\mathrm{softplus}(Wx+b)$ are convex in $x$. Consequently, the feed-forward stack remains convex in its inputs. When paired with the convex softplus--logistic loss (with $\gamma=0$), the first stage behaves as a convex warm start, providing a curvature-regularized initialization for the non-convex stages that follow.
\end{remark}

\begin{remark}
	Although the convex warm start employs convex surrogates (including softplus activations, linear parameter updates, and convex losses), the LSTM layer itself remains fundamentally non-convex, as indicated in Algorithm~\ref{alg:class_imbalance}. This is due to the gating mechanism, which introduces multiplicative interactions between parameters, for example
	\[
	f_t \odot c_{t-1}
	\qquad\text{and}\qquad
	i_t \odot \tilde{c}_t,
	\]
	where $f_t$ and $i_t$ are sigmoid gates and $\tilde{c}_t$ is the candidate cell update. The recurrent structure yields nested, time-dependent non-linear compositions that violate global convexity. The sigmoid function also contains both convex and concave regions, further complicating the curvature. Thus, while the convex warm start confines the optimisation trajectory to a locally well-conditioned region in logit space, the LSTM mapping remains intrinsically non-convex. In this work, we do not attempt to make the overall problem convex; rather, we introduce a convex surrogate in the probability/logit space to provide a well-conditioned initial phase of training.
\end{remark}

\begin{algorithm}[H]
	\small
	\caption{Three-Stage Focal Loss Framework for Class Imbalance with SHAP-Based Explanation}
	\label{alg:class_imbalance}
	
	\KwIn{Training data $\mathcal{D}=\{(x_i,y_i)\}_{i=1}^N$, total epochs $E$, convex cutoff $E_1$, intermediate cutoff $E_2$, fixed $\gamma$}
	\KwOut{Trained LSTM parameters $\theta$ and SHAP-based interpretation}
	
	Initialise LSTM model parameters $\theta$ with convex-aware dense layers\;
	
	\For{$\text{epoch}=1$ \KwTo $E$}{
		\ForEach{batch $(x,y)$ in $\mathcal{D}$}{
			Compute prediction $p=\mathrm{LSTM}(x;\theta)$\;
			
			\uIf{$\text{epoch}\le E_1$}{
				\textbf{Stage 1 (Convex surrogate loss):} compute loss via Algorithm~\ref{alg:convex_focal}\;
			}
			\uElseIf{$E_1<\text{epoch}\le E_2$}{
				\textbf{Stage 2 (Intermediate non-convex focal loss):}\;
				\[
				\mathcal{L}_{\text{total}}
				= -\frac{\alpha_t}{2}(1-p_t)^{\gamma}\log(p_t)
				\]
			}
			\uElse{
				\textbf{Stage 3 (Standard focal loss):}\;
				\[
				\mathcal{L}_{\text{total}}
				= -\alpha_t(1-p_t)^{\gamma}\log(p_t)
				\]
			}
			
			Backpropagate $\mathcal{L}_{\text{total}}$ and update $\theta$\;
		}
	}
	
	\BlankLine
	\textbf{Model Explanation with SHAP}\;
	Select a background dataset $\mathcal{D}_{\text{bg}}\subset\mathcal{D}$\;
	Compute SHAP values $\mathrm{SHAP}(x;\theta,\mathcal{D}_{\text{bg}})$ for selected inputs\;
	Visualise SHAP summary plots and feature attributions\;
	
	\Return{$\theta$ and SHAP-based interpretation}\;
\end{algorithm}

\section{Evaluation}
\label{sec:Evaluation}

\subsection{Dataset Description}
\noindent
The empirical analysis uses a dataset of automobile insurance claims from a large U.S. insurer. The sample contains 39{,}981 records, each associated with a single customer, and 39 explanatory variables describing policy, vehicle, customer, and claim characteristics. These include demographics (e.g., age, gender, education, employment status), policy attributes (coverage type, deductible, premium measures), vehicle features (class, size, year), and claim-related information (total claim amount, reason for filing, and fraud indicator). Key fields such as \texttt{CustomerID}, \texttt{State}, \texttt{Income}, \texttt{Monthly\_Premium\_Auto}, \texttt{Total\_Claim\_Amount}, \texttt{Response}, \texttt{Claim\_Reason}, and \texttt{Coverage} are summarised in Table~\ref{tab:feature-index}.\footnote{\url{https://zenodo.org/records/13381118}}

\begin{figure}[t]
	\centering
	\includegraphics[width=0.85\textwidth]{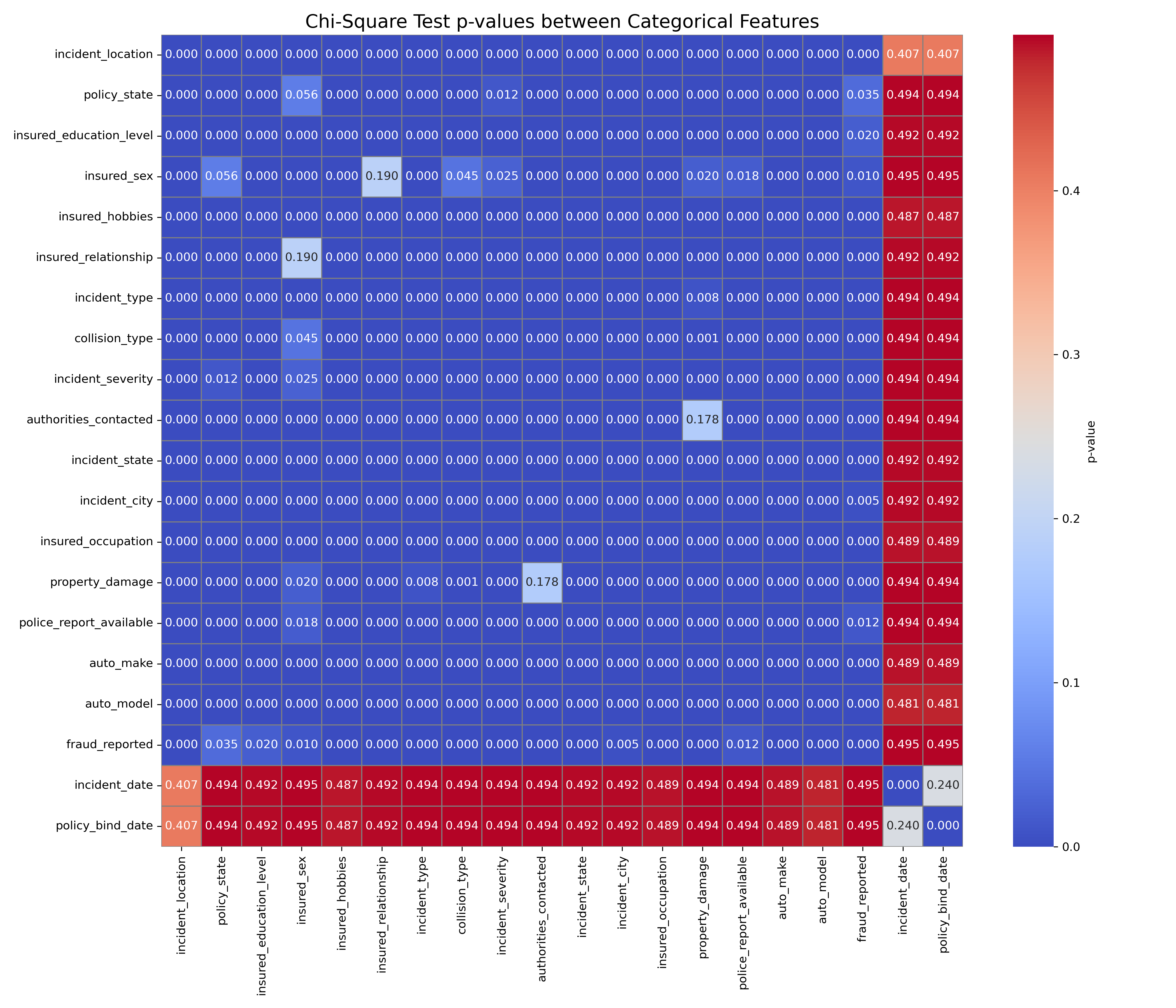}
	\caption{Chi-square test $p$-values for pairwise associations between categorical features in the insurance fraud dataset. Darker blue indicates stronger dependence (small $p$-values), while lighter red denotes weaker association.}
	\label{fig:chisquare}
\end{figure}

\noindent
Figure~\ref{fig:chisquare} displays a chi-square heatmap of pairwise $p$-values for the categorical variables. Many feature pairs exhibit very small $p$-values, suggesting strong dependence. In particular, variables such as \texttt{incident\_location}, \texttt{incident\_type}, \texttt{collision\_type}, \texttt{incident\_state}, \texttt{incident\_severity}, and \texttt{authorities\_contacted} show pronounced association with \texttt{fraud\_reported}, indicating that they are informative for distinguishing fraudulent from non-fraudulent claims. Such variables are naturally important inputs for supervised learning models.
\\
By contrast, some feature pairs appear largely unrelated. For example, \texttt{insured\_sex} versus \texttt{fraud\_reported}, \texttt{policy\_state} versus \texttt{policy\_bind\_date}, and \texttt{insured\_education\_level} versus \texttt{insured\_hobbies} exhibit higher $p$-values (around $0.45$ or above), suggesting only weak dependence. While these variables may still contribute useful marginal information, they add limited incremental value in terms of pairwise interaction. Date-related features such as \texttt{incident\_date} and \texttt{policy\_bind\_date} show moderate $p$-values (around $0.24$), hinting at possible temporal structure in claims and fraud incidence.
\\
To address multicollinearity and improve interpretability, we apply the Variance Inflation Factor (VIF) for feature selection. VIF measures how well a given feature can be explained by the remaining features. Values exceeding 10 are typically viewed as indicative of problematic collinearity. Starting from 39 variables, the VIF analysis reveals substantial redundancy among features such as \texttt{policy\_bind\_date}, \texttt{total\_claim\_amount}, \texttt{auto\_year}, \texttt{vehicle\_claim}, and \texttt{age}. We also remove \texttt{insured\_zip}, \texttt{policy\_csl}, \texttt{policy\_annual\_premium}, and \texttt{number\_of\_vehicles\_involved}. After dropping nine highly collinear variables, we retain a more parsimonious set of 30 features, including policy attributes (e.g., policy state, deductible, umbrella limit), customer demographics (education level, occupation, hobbies), incident descriptors (incident type, collision type, severity, authorities contacted), and claim components (such as injury claim). The target variable \texttt{fraud\_reported} is used as the response in supervised learning.
\\
By pruning redundant features via VIF, we reduce multicollinearity, simplify the model, and support more stable estimation. This cleaned feature set forms the basis for the downstream modelling and evaluation of the proposed three-stage focal-loss framework.
\begin{figure}[H]
	\centering
	\includegraphics[width=0.7\textwidth]{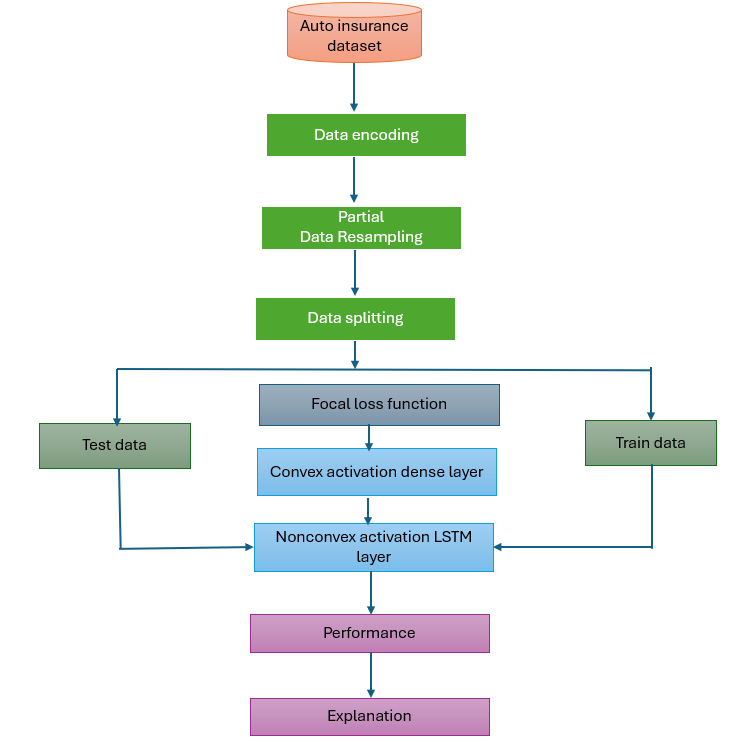}
	\caption{Overview of the auto-insurance fraud detection system under class imbalance, with three-stage focal-loss training and SHAP-based model explanation.}
	\label{fig:system}
\end{figure}

\subsection{Results}
\noindent
All experiments were conducted on a MacBook Pro (2019, 16 GB RAM, Intel Core i7) running macOS. The models were implemented in Python 3.9 using TensorFlow 2.13. Batch sizes and early-stopping criteria were tuned to ensure efficient training given the hardware constraints.
\\
We adopted a 10-fold cross-validation scheme. The original dataset was partitioned into ten equally sized folds; in each iteration, nine folds were used for training and the remaining fold for validation. This procedure was repeated until each fold had served once as the validation set. The reported performance metrics are averages over the ten runs. Such a protocol is particularly appropriate for imbalanced classification problems, where a single train–test split may yield misleading conclusions due to the rarity of the minority class. To address class imbalance, we employed a hybrid resampling strategy combining undersampling of the majority class with oversampling of the minority class. In particular, we applied partial resampling at the 80\% level to improve class balance while preserving the overall distributional structure. This approach reduces the risk that the model becomes biased toward the non-fraudulent class, while limiting the extent to which synthetic examples or discarded majority cases distort the underlying data-generating process.
\\
During training, we used the proposed three-stage focal-loss framework. Relative to standard cross-entropy, focal loss down-weights easy-to-classify examples and concentrates learning on harder, misclassified instances, a property that is especially valuable in imbalanced settings. In our setup, the loss-level adjustment complements the resampling strategy, ensuring that class imbalance is addressed both at the data level and within the optimization objective. We fixed the total number of training epochs at 100 for all methods. For the three-stage approach, the first 10 epochs were allocated to the convex stage (softplus activation, $\gamma = 0$, $\alpha = 0.1$), followed by 40 epochs in the intermediate non-convex stage and 50 epochs in the standard non-convex focal-loss stage. The focusing parameter was set to $\gamma = 4$ in the non-convex stages. A learning rate of $0.001$ was used throughout, with a dense layer of 32 units and an LSTM layer of 128 units across all experimental configurations. The three-stage procedure thus begins with a convex surrogate loss, providing a well-conditioned “warm start”, and then transitions to increasingly expressive non-convex focal-loss phases. The convex initialisation stabilises early optimisation by smoothing the loss surface with respect to the predicted probabilities and reducing sensitivity to local irregularities. Subsequent stages shift emphasis toward harder-to-classify observations by progressively down-weighting easier examples, encouraging the model to refine its decision boundary in regions where the minority class is most difficult to separate.
\begin{table}[H]
	\small
	\centering
	\renewcommand{\arraystretch}{0.5}
	\resizebox{0.5\textwidth}{!}{%
		\begin{tabular}{|c|l|}
			\hline
			\textbf{Index} & \textbf{Feature Name} \\
			\hline
			0  & months\_as\_customer \\
			1  & age \\
			2  & policy\_number \\
			3  & policy\_bind\_date \\
			4  & policy\_state \\
			5  & policy\_csl \\
			6  & policy\_deductable \\
			7  & policy\_annual\_premium \\
			8  & umbrella\_limit \\
			9  & insured\_zip \\
			10 & insured\_sex \\
			11 & insured\_education\_level \\
			12 & insured\_occupation \\
			13 & insured\_hobbies \\
			14 & insured\_relationship \\
			15 & capital-gains \\
			16 & capital-loss \\
			17 & incident\_date \\
			18 & incident\_type \\
			19 & collision\_type \\
			20 & incident\_severity \\
			21 & authorities\_contacted \\
			22 & incident\_state \\
			23 & incident\_city \\
			24 & incident\_location \\
			25 & incident\_hour\_of\_the\_day \\
			26 & number\_of\_vehicles\_involved \\
			27 & property\_damage \\
			28 & bodily\_injuries \\
			29 & witnesses \\
			30 & police\_report\_available \\
			31 & total\_claim\_amount \\
			32 & injury\_claim \\
			33 & property\_claim \\
			34 & vehicle\_claim \\
			35 & auto\_make \\
			36 & auto\_model \\
			37 & auto\_year \\
			38 & fraud\_reported \\
			\hline
		\end{tabular}
	}
	\caption{Feature index for the auto-insurance fraud detection dataset.}
	\label{tab:feature-index}
\end{table}

\begin{table}[H]
	\centering
	\caption{Performance comparison across different training schedules.}
	\label{tab:schedule_performance}
	\renewcommand{\arraystretch}{1}
	\resizebox{1\textwidth}{!}{%
		\begin{tabular}{lcccccc}
			\toprule
			\textbf{Schedule} & \textbf{Loss} & \textbf{Accuracy} & \textbf{Precision} & \textbf{Recall} & \textbf{F1} & \textbf{AUC} \\
			\midrule
			Convex $\alpha=0.1$     & 0.069232 & 0.573024 & 0.668889 & 0.085478 & 0.144526 & 0.531695 \\
			Three Stage             & 0.022523 & 0.606905 & 0.587640 & 0.332631 & 0.414865 & 0.630347 \\
			Nonconvex $\alpha=0.25$ & 0.017794 & 0.559038 & 0.205000 & 0.029418 & 0.050840 & 0.586044 \\
			Nonconvex $\alpha=0.5$  & 0.021273 & 0.568897 & 0.339423 & 0.161636 & 0.205153 & 0.580044 \\
			\bottomrule
		\end{tabular}
	}
\end{table}
\noindent
Table~\ref{tab:schedule_performance} summarizes the performance of the four training schedules. The three-stage scheme achieves the highest F1 score and AUC, together with the best overall accuracy among the compared methods. Its recall is substantially higher than that of the alternatives, indicating improved detection of fraudulent cases while maintaining a competitive level of precision. By contrast, the convex-only schedule yields the highest precision but extremely low recall, resulting in a poor F1 score. This pattern suggests that the convex model is overly conservative: it correctly identifies many of the flagged cases but fails to recognise a large portion of fraudulent observations. The non-convex schedule with $\alpha = 0.25$ attains the lowest loss value but exhibits very weak recall and F1, indicating a misalignment between the training objective and the effective separation of the minority class. Increasing $\alpha$ to $0.5$ improves recall and F1 relative to the $\alpha = 0.25$ configuration, yet performance still falls short of the three-stage approach in terms of AUC and balanced detection.
\\
Overall, the three-stage schedule appears to offer the best trade-off across the metrics, combining the stable initial behaviour of convex training with the representational flexibility of non-convex refinement.
\begin{figure}[H]
	\centering
	\includegraphics[width=0.9\textwidth]{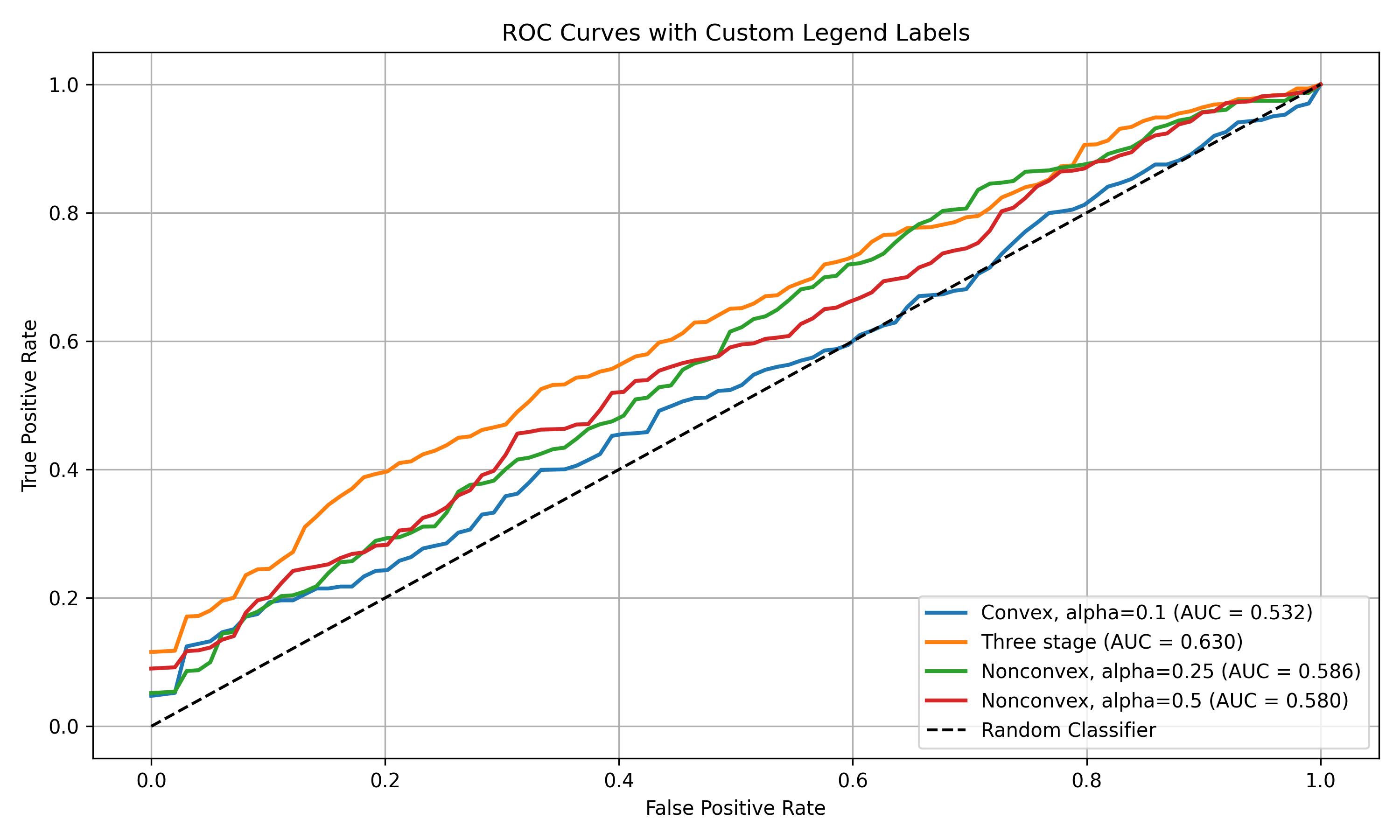}
	\caption{ROC curves for the different training schedules under class imbalance.}
	\label{fig:roc_mean}
\end{figure}
\noindent
The ROC curves in Figure~\ref{fig:roc_mean} provide a complementary view of these results. The three-stage procedure achieves the highest AUC, consistent with the tabulated metrics. From an optimisation perspective, this suggests that a gradual transition from a convex objective to a non-convex focal loss effectively guides the optimiser through a smoother loss landscape in the early stages, before introducing additional curvature and expressiveness. In this sense, the staged procedure functions as an implicit continuation method, reducing the likelihood of early convergence to poor local minima. In contrast, the purely non-convex models operate on a highly rugged objective from the outset, making them more sensitive to initial conditions and more prone to convergence in suboptimal regions. The convex-only model, while well-conditioned from an optimisation standpoint, lacks sufficient flexibility to represent complex decision boundaries, leading to underfitting and weaker discrimination. These observations indicate that neither purely convex nor purely non-convex optimisation is optimal in isolation; instead, combining convex initialisation with non-convex refinement yields more favourable convergence and generalisation properties in this imbalanced classification setting.

\begin{figure}[H]
	\centering
	\includegraphics[width=0.9\textwidth]{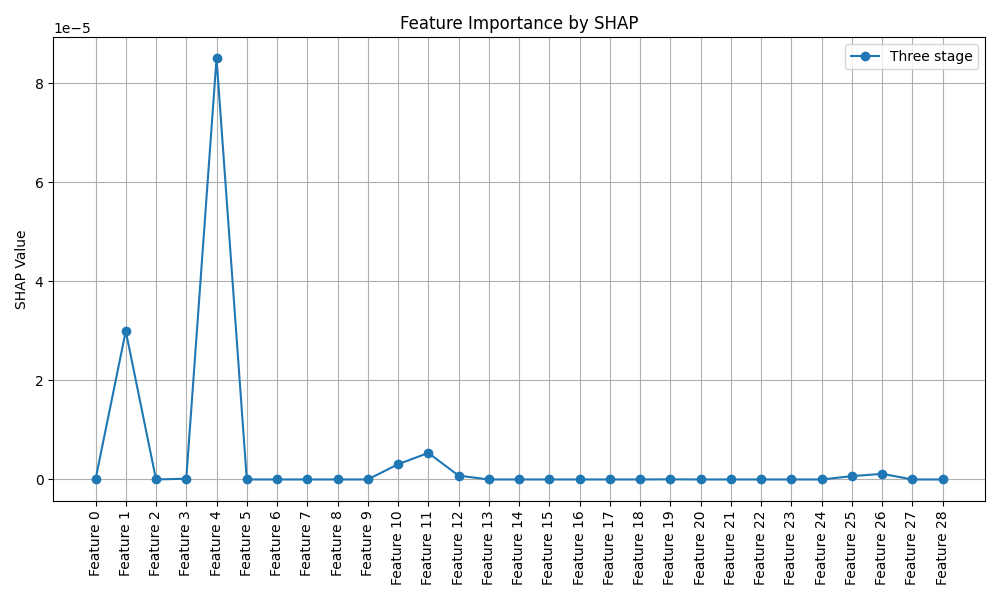}
	\caption{Feature index vs.\ SHAP values under class imbalance.}
	\label{fig:shap_label}
\end{figure}
\noindent
Figure~\ref{fig:shap_label} shows the SHAP-based feature-importance profile for the three-stage model. The attribution pattern is relatively sparse, with \texttt{policy\_state} (Feature~4) emerging as the most influential predictor, followed by \texttt{age} (Feature~1). Smaller but non-negligible contributions arise from variables such as \texttt{insured\_sex} and \texttt{insured\_education\_level} (Features~10–11), while several other features exhibit only minor SHAP magnitudes. This concentration of importance on a small set of variables is consistent with domain knowledge: geographic and demographic characteristics often play a central role in fraud risk assessment. The corresponding feature indices and descriptions are summarised in Table~\ref{tab:feature-index}.
\\
It is worth noting that the SHAP profile is influenced by the use of partial SMOTE-based oversampling (Figure~\ref{fig:system}). Oversampling reshapes the local decision boundary around minority-class observations and can redistribute feature influence across the fraud manifold. As a result, while the dominant drivers remain broadly stable, the relative magnitudes of secondary features may vary under alternative resampling schemes. This underscores that the explanations reflect both the model structure and the data-preprocessing choices, rather than being purely intrinsic to the model alone.

\section{Conclusion and Recommendation}
\label{sec:Conclusion}
\noindent
This study compared four training schedules for auto-insurance fraud detection under severe class imbalance: a convex-only scheme with softplus activation, two purely non-convex focal-loss configurations (with $\alpha = 0.25$ and $\alpha = 0.5$), and a proposed three-stage convex-to-non-convex strategy. Using accuracy, precision, recall, F1 score, and AUC as evaluation metrics, we found that the three-stage schedule delivers the most balanced and reliable performance. By combining the stable optimisation behaviour of a convex warm start with the greater expressiveness of non-convex refinement, the three-stage approach outperforms both convex-only and non-convex-only baselines on nearly all metrics. The three-stage framework is particularly effective in addressing class imbalance. It captures complex fraud patterns associated with rare events while maintaining a competitive AUC relative to the convex-only model. In practice, this results in a more favourable trade-off between sensitivity, generalization, and stability, highlighting the robustness of the proposed training schedule for real-world auto insurance fraud detection systems operating under extreme skew. We also examined feature-level behaviour via SHAP-based attributions. The staged transition from convex to non-convex learning led to selective and well-regularized feature utilization, producing a sparse and structured attribution profile in which only a small subset of discriminative variables carried substantial weight. Importantly, even the most influential features exhibited moderate, bounded SHAP magnitudes, suggesting that the three-stage schedule achieves an effective balance between model expressivity and regularization. This indicates that the staged optimization not only improves predictive accuracy but also stabilizes the internal decision structure by avoiding both diffuse underuse of relevant features and brittle over-concentration on a few inputs.
\\
We recommend that future studies should focus on adaptive mechanisms for automatically determining the transition points between convex and non-convex phases, as well as extending the proposed framework to other high-impact domains such as cybersecurity, mission-critical operational analytics, and broader financial fraud detection, where imbalanced data and subtle feature interactions continue to pose significant modelling challenges.

	\section*{Acknowledgements}
This research has been supported in part by the Faculty of Science at Toronto Metropolitan University.
	
	

\end{document}